\documentclass[conference]{IEEEtran}
\IEEEoverridecommandlockouts
\usepackage{cite}
\usepackage{amsmath,amssymb,amsfonts}
\usepackage{algorithmic}
\usepackage{graphicx}
\usepackage{textcomp}
\usepackage{xcolor}
\def\BibTeX{{\rm B\kern-.05em{\sc i\kern-.025em b}\kern-.08em
    T\kern-.1667em\lower.7ex\hbox{E}\kern-.125emX}}

\usepackage{amsmath}
\usepackage{array}
\usepackage{multirow}
\usepackage{colortbl}
\usepackage{subcaption}
\usepackage{titlesec}
\usepackage{tikz}
\usepackage[linesnumbered,ruled,vlined,boxed]{algorithm2e}
\usepackage{wrapfig}
\usepackage{graphicx}
\usepackage{caption}
\usepackage{geometry}
\usepackage{booktabs}

\definecolor{customgreen}{HTML}{00C474} 
\definecolor{customred}{HTML}{ff6768}   

\newcommand{\customgreenuparrow}{
    \begin{tikzpicture}
    \draw[customgreen, -latex] (0,0) -- (0,0.35);
    \end{tikzpicture}
}

\newcommand{\customreddownarrow}{
    \begin{tikzpicture}
    \draw[customred, -latex] (0,0) -- (0,-0.35);
    \end{tikzpicture}
}

\definecolor{algref_color}{HTML}{5E81B5} 
\newcommand{\algcomm}[1]{\textcolor{algref_color}{$\triangleright$ #1}}

\definecolor{ourwork_clr}{HTML}{6495ED}
\newcommand{\temtextclr}[1]{\textcolor{ourwork_clr!30}{\rule{#1}{10pt}}}

\begin{document}
\title{Rehearsal-free Federated Domain-incremental Learning
}


\author{Rui Sun$^{\dagger}$, Haoran Duan$^{*\dagger}$, Jiahua Dong$^{*\ddagger}$, Varun Ojha$^{\dagger}$, Tejal Shah$^{\dagger}$, and Rajiv Ranjan$^{\dagger}$ \\
$^{\dagger}$ Newcastle University, Newcastle upon Tyne, UK \\
$^{\ddagger}$ Mohamed bin Zayed University of Artificial Intelligence, Abu Dhabi, UAE \\
$^{\dagger}$ \{rui.sun, haoran.duan, varun.ojha, tejal.shah, raj.ranjan\}@newcastle.ac.uk \\ 
$^{\ddagger}$ dongjiahua1995@gmail.com

\IEEEcompsocitemizethanks{\IEEEcompsocthanksitem * These authors are the corresponding authors.}
}

\maketitle

\begin{abstract}
We introduce a rehearsal-free federated domain incremental learning framework, RefFiL, based on a global prompt-sharing paradigm to alleviate catastrophic forgetting challenges in federated domain-incremental learning, where unseen domains are continually learned. Typical methods for mitigating forgetting, such as the use of additional datasets and the retention of private data from earlier tasks, are not viable in federated learning (FL) due to devices' limited resources. Our method, RefFiL, addresses this by learning domain-invariant knowledge and incorporating various domain-specific prompts from the domains represented by different FL participants. A key feature of RefFiL is the generation of local fine-grained prompts by our domain adaptive prompt generator, which effectively learns from local domain knowledge while maintaining distinctive boundaries on a global scale. We also introduce a domain-specific prompt contrastive learning loss that differentiates between locally generated prompts and those from other domains, enhancing RefFiL's precision and effectiveness. Compared to existing methods, RefFiL significantly alleviates catastrophic forgetting without requiring extra memory space, making it ideal for privacy-sensitive and resource-constrained devices.
\end{abstract}


\section{Introduction}
Federated Learning~(FL) is a remarkable approach that enables various participants, also referred to as clients, to collaboratively train a global machine learning model while keeping their data decentralized, maintaining privacy and leveraging diverse data sources~\cite{mcmahan2017communication}. A significant limitation of current FL research lies in its primary focus on scenarios with 
static data distributions~\cite{li2020federated, karimireddy2020scaffold}, which remain unchanged throughout the training process~\cite{dong2022federated,dong2023no}. This limitation is particularly pronounced, as it fails to reflect the dynamic nature of real-world data. Recent studies have therefore shifted towards Federated Continual Learning (FCL), especially in class-incremental data scenarios~\cite{dong2022federated,dong2023no,zhang2023target,ma2022continual,babakniya2024data}, which struggles significantly with the critical problem of \textit{catastrophic forgetting}~\cite{rebuffi2017icarl}, where model rapidly forgets previously learned knowledge upon encountering new data. In this paper, we explore more challenging domain-incremental data scenarios.

Various methods have been proposed to address catastrophic forgetting. Rehearsal-based approaches, as described by Smith et al.~\cite{rebuffi2017icarl}, mitigate forgetting by replaying a selection of old task data during new task training. Alternatively, regularization-based methods, like Knowledge Distillation 
(KD)~\cite{li2017learning,rebuffi2017icarl,dong2022federated}, guide the model's optimization using historical data or models. Network expansion techniques~\cite{kirkpatrick2017overcoming, ermis2022memory, pham2021dualnet} adapt the model structure to incorporate new features but their practicality in memory-constrained FL environments is limited. Recent research like L2P~\cite{wang2022learning} and DualPrompt~\cite{wang2022dualprompt} explores prompt-based, rehearsal-free domain-incremental learning. However, these methods still do not fully address the challenge of using global domain information to enhance local model robustness in Federated Domain-Incremental Learning~(FDIL).


Strategies for mitigating catastrophic forgetting in domain-incremental tasks differ from those employed in class-incremental scenarios, where substantial differences between classes enable models to form clearer decision boundaries. In contrast, domain-incremental contexts involve more subtle inter-domain variations, which challenge a model’s ability to effectively separate domains. As a result, the focus shifts toward acquiring domain-invariant knowledge—an essential step in enhancing model robustness. This becomes particularly critical when models are expected to generalize across a broad spectrum of diverse domains. Despite its importance, research in this area remains limited, particularly regarding the global utilization of heterogeneous domain information from all clients throughout their learning progression. Such utilization is crucial for mitigating catastrophic forgetting by improving local models’ ability to learn domain-invariant representations in FDIL and, more broadly, within the FCL paradigm.

\begin{figure*}[th]
  \centering
  \includegraphics[width=\linewidth]{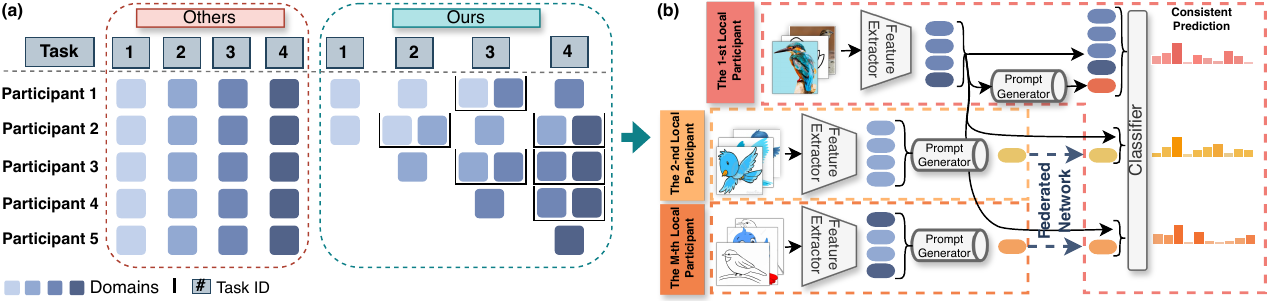}
  \caption{Key steps of RefFiL framework. Left panel (a) shows the common setting in existing FCL works, characterized by a cliff-style data transition. Right panel (a) shows our approach, where a subset of participants gradually transitions to new tasks. Panel (b) provides an overview example of a key step in our methodology: the 1st participant processes new domain data using global prompts from the 2nd to $m$-th participants and local prompts, enhancing robustness by aligning the model's predictions across diverse domain prompts as inputs.}
  \label{fig:into_overall_scenario}
\end{figure*}

In this paper, we introduce \textbf{RefFiL}, a \textit{rehearsal-free federated domain-incremental learning} framework that harnesses a wide spectrum of global domain information to augment cross-domain prediction robustness by specifically enhancing the acquisition of domain-invariant knowledge. We first recognize the inherent challenges posed by the high autonomy of participants in FL, which introduces significant uncertainty in data distribution and, consequently, a substantial domain gap. Thus, derivating robust prompts becomes critical in learning domain-invariant knowledge. 

Hence, firstly, we define a practical FDIL problem. To address the challenge of significant domain gaps in datasets from various resource-sensitive participants, we propose a \textit{client-wise domain adaptive prompt generator}, which creates personalized, instance-level prompts for each participant's data, incorporating domain adaptation knowledge for global sharing. Then, to effectively break down the silos of domain-specific information in FL, we introduce a \textit{global prompt learning paradigm}. This paradigm involves globally distributing and sharing clustered local prompts combined with local domain-invariant knowledge learning through prompting. By doing so, we aim to empower the local model to effectively differentiate between prompt features from various domains thereby enhancing its ability to learn domain-invariant knowledge from local data, which is crucial for boosting the model's robustness in multi-domain predictions and significantly reducing its susceptibility to catastrophic forgetting. Finally, we introduce a \textit{novel domain-specific prompt contrastive learning} method with temperature decay aimed at improving the local model's ability to distinguish between semantically related and unrelated prompts. By seamlessly integrating these three key components, the RefFiL framework emerges as a highly competitive method in the realm of FDIL.

\section{Preliminaries}

\noindent{\textbf{Federated Domain-incremental Learning.}}
In standard domain-incremental learning~\cite{mirza2022efficient}, a series of sequential tasks are denoted as $\mathcal{T} = \{\mathcal{T}^t\}_{t=1}^T$, with $T$ indicating the total number of tasks. Each task $\mathcal{T}^t$, the $t$-th in the sequence, consists of $N^t$ data-label pairs, i.e., $\mathcal{T}^t=\left\{x_i^t, y_i^t\right\}_{i=1}^{N^t}$. Although each task is associated with a distinct domain \(\mathcal{D}^t\), all tasks share the same label space \(\mathcal{Y}\). The complete domain space is defined as \(\mathcal{D} = \{\mathcal{D}^t\}_{t=1}^T\), with task domains differing in data distribution, i.e., \(\mathcal{D}^t \neq \mathcal{D}^{t-1}\).

We adapt this setting to the FDIL scenario. Here, $M$ local participants, denoted as $\mathcal{M} = \left\{\mathcal{M}_m\right\}_{m=1}^M$, interact with a central global server $\mathcal{M}_G$. During each global communication round $r$ (where $r = 1, \dots, R$), a subset of local participants is randomly selected for gradient aggregation. When a participant $\mathcal{M}_l$ is selected for the $t$-th incremental task, it receives the latest global model $\theta^{{r},t}$.

The participant then trains this model on its local dataset $\mathbf{D}^t \sim \mathcal{P}_l^{\left|\mathbf{D}^t\right|}$, specific to task $t$, where $\mathcal{T}_m^t=\left\{x_{m i}^t, y_{m i}^t\right\}_{i=1}^{N_l^t} \subset \mathcal{T}^t$ consists of data from new domains, and $\mathbf{P}_m$ denotes the data quantity distribution of participant $m$. These local datasets $\left\{\mathbf{P}_m\right\}_{m=1}^M$ are non-independent and identically distributed (non-iid), exhibiting a form of quantity shift. Each incremental task includes $K^t$ classes, maintaining consistent class counts ($K^1 = K^2 = \dots = K^t$) and labels ($\mathbf{y}^1 = \mathbf{y}^2 = \dots = \mathbf{y}^t$), but domain and data distributions vary across tasks ($\mathbf{x}^1 \neq \mathbf{x}^2 \neq \dots \neq \mathbf{x}^t$). Once $\mathcal{M}_m$ completes training on task $t$ with model $\theta^{r,t}$, it obtains an updated model $\theta^{r,t}_l$. All selected participants transmit their updates to the global server $\mathcal{M}_G$, where aggregation produces the updated global model $\theta^{(r+1),t}$. This model is then broadcast back to all participants, initiating the next communication round.

\noindent{\textbf{Client increment strategy.}}
We divide participants into three dynamic groups for each incremental task: Old $\mathcal{U}_o$, In-between $\mathcal{U}_b$, and New $\mathcal{U}_n$. $\mathcal{U}_o$, with $M_o$ members, accesses only past domain data. $\mathcal{U}_b$ (with $M_b$ clients) works on both new and old domain data, while $\mathcal{U}_n$, consisting of $M_n$ new participants, focuses solely on new domain data from the current task. These groupings are redefined at each global round, with $\mathcal{U}_n$ being irregularly introduced during the FDIL process. As tasks progress, this strategy gradually increases the total number of participants, $M = M_o + M_b + M_n$.

\noindent{\textbf{Learning with Prompts.}}
Prompt techniques, originally developed for task adaptation in natural language processing~\cite{liu2023pre}, have been primarily applied to large transformer models. However, they can also be adapted to convolutional neural network (CNN) based architectures, making them suitable for resource-constrained federated learning environments.

In our case, the model backbone includes a CNN-based feature extractor $h$, a patch embedding layer, attention block $b$, and a classifier $G$. Given an input image $x$, the feature map is obtained as $F = h(x)$. This is split into $n$ $d$-dimensional patch tokens $\mathrm{PT}$, and a trainable [CLS] token is prepended, forming the input token sequence:
\begin{equation}
\mathbf{I} =\left[\mathrm{CLS} ; \mathrm{PT}_1, \dots, \mathrm{PT}_n \right] \in \mathbb{R}^{(n+1) \times d}.
\end{equation}
This token sequence is then fed into the $b$-th attention block, producing:
\begin{equation}
\mathbf{I}_{b+1}=\operatorname{LN}\left(\mathbf{I}_b^{\prime}+\mathbf{I}_b^{\prime \prime}\right),
\end{equation}
where LN is layer normalization, $\mathbf{I}_b^{\prime}=\operatorname{LN}(\operatorname{MHSA}(\mathbf{I}_b))$ applies multi-head self-attention (MHSA), and $\mathbf{I}_b^{\prime \prime}=\operatorname{MLP}(\mathbf{I}_b^{\prime})$ is a multilayer perceptron (MLP). Each attention block consists of an MHSA layer, followed by MLP, skip connections~\cite{he2016deep}, and LN~\cite{ba2016layer} (see Fig.~\ref{fig:main_stru}). The classifier $G$ maps the final [CLS] token to the class prediction:
\begin{equation}
\hat{y} = G([\mathrm{CLS}]_B),
\end{equation}
where $\hat{y}$ denotes the predicted class probability distribution.

\section{Related Work}
Most recently, several works have begun to concentrate on FCL, especially Federated Class-incremental Learning~(FCIL), which involves sequentially learning a series of distinct, non-overlapping classes. The primary challenge in FCIL is catastrophic forgetting, where models tend to lose previously learned knowledge upon task transition. Approaches such as GFLC~\cite{dong2022federated} and its extension LGA~\cite{dong2023no} use knowledge distillation, employing an old model to regularize the current model's output and a loss function to mitigate global forgetting. Method TARGET~\cite{zhang2023target} uses a global model to generate synthetic data with a global distribution for the current training task to reduce forgetting. CFed~\cite{ma2022continual} addresses data unavailability by generating pseudo-labels on a surrogate dataset for knowledge distillation and proposes a server distillation mechanism to mitigate intra-task forgetting. These methods typically require storing old tasks or extra data, which is impractical in FL with resource-limited devices. MFCL~\cite{babakniya2024data} utilised GAN to generate data rather than save raw to achieve rehearsal-free learning from class-incremental data. Despite progress in the class-incremental scenario of FCL, there remains a notable gap in the domain-incremental setting, which focuses on learning domain-invariant knowledge to reduce forgetting, a different approach from FCIL.

\section{The Proposed Framework RefFiL}
\label{sec:pre}
\begin{figure*}[ht]
  \centering
  \includegraphics[width=1\textwidth]{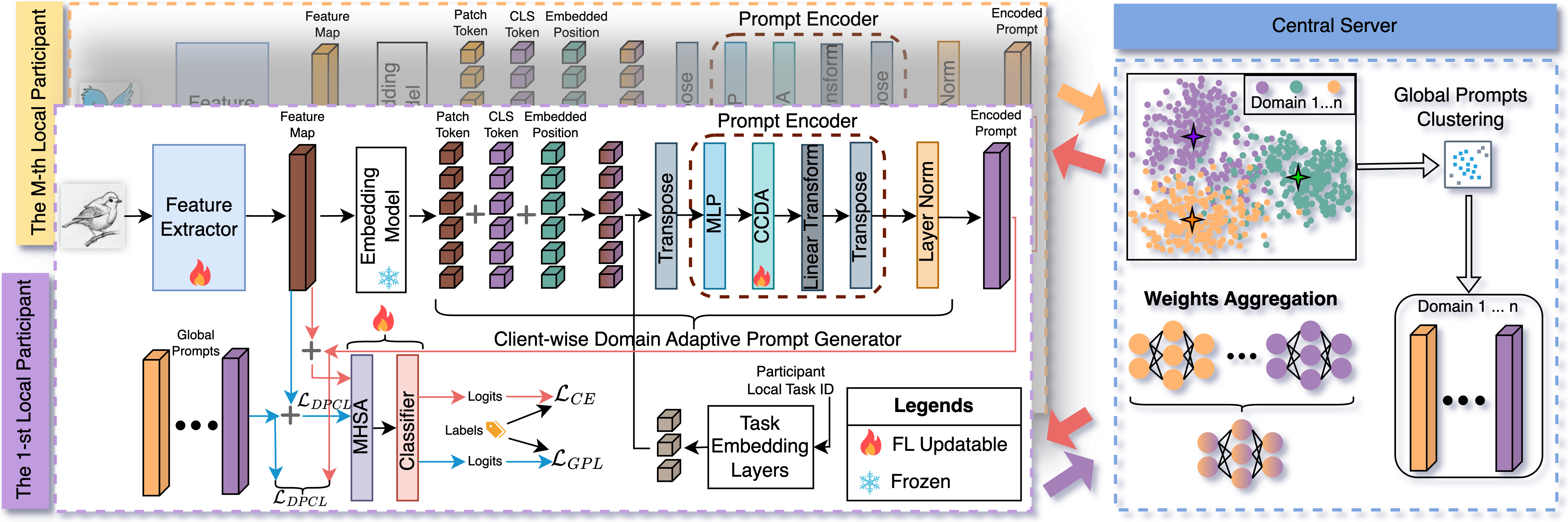}
  \caption{Overview of the RefFiL Framework. Each participant first encodes local prompts using the tokenized feature map and task ID embedding. These local prompts are then concatenated with the feature map to compute the loss $\mathcal{L}_{CE}$. Simultaneously, the feature map is combined with global prompts to calculate the loss $\mathcal{L}_{GPL}$, and the loss $\mathcal{L}_{DPCL}$ is determined between global and local prompts. Subsequently, all local prompts, along with the updated local models, are transmitted to the central server. The server then clusters the prompts domain-wise and aggregates the local models for distribution in the next training round.}
  \label{fig:main_stru}
\end{figure*}

To address the devices' resource-constrained limitation in FDIL, the concept of prompt techniques, initially introduced in natural language processing (NLP) for task adaptation~\cite{liu2023pre}, offers a promising solution. These techniques append instructions to large pre-trained models, helping them use existing knowledge instead of learning from scratch. Applied to continual learning, approaches like L2P~\cite{wang2022learning} and DualPrompt~\cite{wang2022dualprompt} use prompts to avoid the need for rehearsal. L2P uses a prompt pool for task-specific knowledge, replacing the rehearsal buffer. DualPrompt, designed for resource-limited devices, uses two prompt types: General-Prompt for general instructions and Expert-Prompt for task-specific guidance to replace maintaining a prompt pool. However, in FDIL, the challenge goes beyond learning local domain-invariant information; it involves using data from other participants to enhance model robustness against diverse domains. This paper aims to address this specific challenge, unexplored in current research.

To enhance local domain-invariant knowledge learning through the sharing of diverse domain information via global prompts in FDIL settings, we introduce RefFiL, a rehearsal-free framework that leverages global prompts to share diverse domain information. As shown in Fig.~\ref{fig:main_stru}, RefFiL operates in task stages where participants engage in different domains. The process starts with a feature extractor generating a feature map from the input image, followed by our Client-wise Domain Adaptive Prompt~(CDAP) generator. This generator encodes prompts using the feature map with the task ID. These encoded prompts, combined with the feature map, undergo classification involving computation of the cross-entropy loss and the Domain-specific Prompt Contrastive Learning (DPCL) loss with global prompts. Additionally, the feature map is merged with global prompts to compute the Global Prompt Learning (GPL) loss. All local prompts are sent to the central server to facilitate domain information sharing.

\noindent{\textbf{Client-wise Domain Adaptive Prompt Generator.}}
Most of the stages in FDIL training typically involve datasets from various domains, creating a substantial domain gap between participants' local data. To navigate this challenge, generating refined prompts is essential for enabling participants to effectively distinguish and share prompts globally. Common approaches include maintaining a prompt pool, as seen in methods like L2P~\cite{wang2022learning} and DualPrompt~\cite{wang2022dualprompt}. These strategies are similar to traditional continual learning methods, which rely on storing raw data or summarised representative information. However, a significant obstacle in FL is the limited memory resources of participant devices and privacy sensitivity. Additionally, as noted in \cite{jung2023generating}, a fixed-size prompt pool may not effectively encode domain-specific knowledge, and a scalable pool faces feasibility issues due to uncertainties in task requirements and computational constraints. This makes maintaining an extensive and adaptable prompt pool challenging for continuous learning in a federated context, where the unpredictability of tasks and resource limitations of devices must be considered.

Thus, inspired by prompt encoder network structure proposed in~\cite{jung2023generating}, we propose a client-wise domain adaptive prompt generator. This generator is designed to create personalized prompts for each participant's local data, embedding instance-level, fine-grained instructions with domain adaptation knowledge for global sharing. The CDAP generator $\mathcal{G}$ comprises several components: Layer Normalization (LN), a Multi-Layer Perceptron (MLP), a Cross-Client Domain Adaptation (CCDA) layer, and a Linear Transformation Layer (LT). This configuration enables the generation of adaptive local prompts $p_m$ for each participant $m$, which could be formulated as:
\begin{equation}
\label{alg:ldap}
\begin{aligned}
P_m & =\operatorname{LT}\left(\operatorname{CCDA}\left(\operatorname{MLP}\left(\operatorname{LN}(\mathbf{I})^{\top}\right)\right) ; \phi(v)\right)^{\top} \\
& =\left(\alpha_{v} \operatorname{CCDA}\left(\operatorname{MLP}\left(\operatorname{LN}(\mathbf{I})^{\top}\right)+\lambda_{v}\right)^{\top}\right) \in \mathbb{R}^{p \times d} \text{,}
\end{aligned}
\end{equation}
where $\mathbf{I}$ is a set of input tokens, $\phi$ represents a linear layer that predicts two types of affine transformation parameters, $\alpha$ and $\lambda$, from a conditional input embedding $v$. Initially, $v$ is normalized, transposed, from $(n + 1) \times d$ to $d \times (n + 1)$ and then passed through an MLP to generate $p$-dimensional outputs which create current task domain well-fitted and instance-level prompts tailored with domain-specific knowledge. To handle the data that vary significantly in distribution from various domains, we introduce CCDA layer, a globally transferable linear layer enhancing the CDAP generator's generalization and robustness via FL aggregation. Additionally, we incorporate a Feature-Wise Linear Modulation (LT)~\cite{perez2018film} framework to augment the generated prompts with additional instructional guidance for prediction. LT achieves this by applying an affine transformation to the instance-level prompts, using scaling $\alpha_v$ and shifting parameters $\lambda_v$, both conditioned on the embedding $v$ as $[\alpha_v, \lambda_v] = \phi(v)$. For the conditional input embedding $v$, we designed a task-specific key embedding layer that is also employed to encode local task IDs, linking tasks with domain-specific data. However, the task ID contributes only to training process by helping to distinguish data from different tasks, and is not utilized during the inference stage.

\noindent{\textbf{Global Prompt Sharing.}}
To overcome domain information silos in FL and address the challenge of data quantity disparity among participants, we propose globally distributing local prompts in a balanced manner. This method prevents network overload in scenarios involving thousands of devices and mitigates biases due to uneven data distribution. Our solution involves averaging across all clients, ensuring equitable influence from each participant on the global model, regardless of their data volume. This strategy is crucial in situations where differences in data quantity between resource-rich and resource-poor participants could significantly skew the global model optimization direction. The formulation for the Local Prompts Group ($LPG$) is as follows:
\begin{equation}
LPG_m=\left[\frac{1}{N} \sum_{j=1}^N P^1_{m,j} ,\dots, \frac{1}{N} \sum_{j=1}^N P^k_{m,j} \right]\in \mathbb{R}^d \text{,}
\end{equation}
then global prompts set~$P^g$ could be summarised as:
\begin{equation}
P^g = \left[LPG_1,\dots, LPG_m\right] \text{.}
\end{equation}
However, generally, a significant portion of participants, 80\%  in our scenario, transition to training tasks in new domains, while a minority continues with old tasks. This leads to a domain-specific prompt imbalance, with most prompts from new domains. This imbalance may bias local model updates towards these majority domains.

\noindent{\textbf{Global Prompts Clustering.}}
To tackle the prompt imbalance issue, directly averaging all prompts may lead to a loss of important domain-characterized features, as averaging might not fully capture the unique characteristics essential for effective learning in diverse domains. Therefore, we propose the use of global prompts clustering that filters and selects representative prompts for each class based on their semantic domain characteristics. We adopt FINCH~\cite{sarfraz2019efficient}, a parameter-free clustering method and particularly well-suited for large-scale prompts clustering with minimal computing resources, which is FL dynamic environment friendly. Additionally, its efficiency in clustering feature prototypes in FL has been previously demonstrated in~\cite{huang2023rethinking}. FINCH operates on the principle that the nearest neighbor prompt of each sample can effectively support grouping. This is highly effective for distinguishing between prompts from different domains, as prompts from separate domains are unlikely to be nearest neighbors and are thus grouped separately. In contrast, similar prompts, likely from the same domain, are grouped together. 
Prompt similarity is evaluated using cosine similarity, where prompts are grouped by their smallest distance. This allows us to sort prompts into domain-specific sets efficiently. For each class $k$ of prompts, we then construct an adjacency matrix as:
\begin{equation}
\label{alg:cluster_adj_matrix}
A^k(m, j)= \begin{cases}1, & \text { if } j=c_m^k \text { or } m=c_j^k \text { or } c_m^k=c_j^k \\ 0, & \text { otherwise }\end{cases} \text{,}
\end{equation}
where $c_m^k$ denotes the nearest neighbor (with the highest cosine similarity) of the class $k$ prompt from the $m$-th participant. Based on this clustering (Eq.\ref{alg:cluster_adj_matrix}), we select representative prompts for each class from the embedding space. The final global clustered prompts are denoted as $\hat{P}^{g}$, are obtained as:
\begin{equation}
\begin{aligned}
\label{eq:global_clustering}
\hat{P}^{g,k}=\left\{LPG_m\right\}_{m=1}^C  \xrightarrow{\Psi} \left\{LPG_m\right\}_{m=1}^N \in \mathbb{R}^{N \times d} \text{,} \\
\hat{P}^g = \left[\hat{P}^{g,1},\dots, \hat{P}^{g,k}\right] \text{.}
\end{aligned}
\end{equation}
Finally, $C$ prompts are clustered by cluster model $\Psi$ into $N$ representatives of class $k$.

\noindent{\textbf{Personalised Prompt Generator Optimization.}}
We expect to improve the local model's ability to differentiate between semantically related and unrelated prompts, ensuring a clear class decision boundary by enhancing sensitivity to domain-specific nuances in prompts. Specifically, we encourage the generation of prompts that are closely aligned with a semantically similar prompts set ($P^+_i$) while being distinctly different from a semantically unrelated prompts set ($P^-_i$). The prompts sampling strategy is defined as follows: For each class \(k\), global prompts for class \(k\) are selected, and their cosine similarity with the locally generated prompt of \(k\) is computed. For clients \(\mathcal{U}_o\) and \(\mathcal{U}_n\) (single domain), the closest prompt is the positive sample \(P^+\). For clients \(\mathcal{U}_b\) (two domains), the two closest prompts are positive samples, and all remaining prompts are negative samples \(P^-_i\). Drawing inspiration from FPL~\cite{huang2023rethinking}, where contrastive learning was used to augment the model's proficiency in learning domain-invariant knowledge, we introduce domain-specific prompt contrastive learning with temperature decay specifically designed to enhance the adaptability of prompt generation in the CDAP. It is natural to derive following optimization objective term:
\begin{equation}
\label{eq:dpcl_loss}
\scalebox{1}{
\scriptsize
    $\mathcal{L}_{DPCL} = -\log \frac{\exp (\text{sim}(u_i, P^+_i) / \tau^{\prime})}{\exp (\text{sim}(u_i, P^+_i)/ \tau^{\prime}) + \exp (\text{sim}(u_i, P^-_i)/ \tau^{\prime})} \text{.}$
}
\end{equation}
Additionally, we address increasing domain diversity globally by proposing a method to calculate the contrastive temperature, $\tau$, which adaptively adjusts the loss. Initially allowing flexibility in distinguishing between positive and negative prompts, it becomes more stringent as learning progresses. The calculation of the updated temperature is as follows:
\begin{equation}
\label{eq:tem_decay}
    \tau^{\prime} = \max\left(\tau_{\text{min}}, \tau \cdot \left(1 - \left(\gamma + (t - 1) \cdot \beta\right)\right)\right) \text{,}
\end{equation}
where $\tau_{\text{min}}$ is the minimum temperature, $\gamma$ represents the base decay rate of the temperature, and $\beta$ is the rate of increment. Both $\gamma$ and $\beta$ are confined within the interval [0,1], indicating that their values range from 0 to 1, inclusive.

\noindent{\textbf{Local Domain-invariant Knowledge Learning with Prompting.}} 
To address the complexity of representing each class in our global clustering model due to varying cluster numbers, we average across all classes. This strategy allows us to create a generalized prompt~$\bar{P}^G$ that effectively represents global domain information, ensuring a balanced and simplified approach to encapsulating the diversity of each domain, where
\begin{equation}
\label{eq:global_p_avg}
\bar{P}^g = \left[AVG(\hat{P}^{g,1}),\dots,AVG(\hat{P}^{g,k}) \right] \text{.}
\end{equation}

We enhance our model's prediction robustness and mitigate forgetting by using global heterogeneous domain-specific prompts as diverse stimuli. This method helps the model learn domain-invariant knowledge. To support this, we introduce the $GPL$ loss, formulated by developing a CrossEntory~\cite{de2005tutorial}~(CE) mechanism, which is defined by the following equation:
\begin{equation}
\label{eq:gpl_loss}
\mathcal{L}_{GPL}=-\sum_{i=1}^{N} y_i \log \left(\text{softmax}\left(\xi_g\right)\right) \text{,}
\end{equation}
where $\mathcal{L}_{GPL}$ a loss function, sums over $N$ instances, using $y_i$ as the true label and applying softmax to the output of logits $\xi_g = G\left([\bar{P}^g, h(x)]\right)$, classifier $G$ which combines the global prompt $\bar{P}^g$ with feature $h(x)$. Concurrently, locally CDAP generated prompts $P^l$ are integrated into the CE loss function, $\mathcal{L}_{CE}$, to capture local domain-specific features, where $\xi_l = G\left([P^l, h(x)]\right)$, thus reinforcing the model's discriminative capability within local domains. The formulation of this loss function is as follows:
\begin{equation}
\label{eq:ce_loss}
\mathcal{L}_{CE}=-\sum_{i=1}^{N} y_i \log \left(\text{softmax}\left(\xi_l\right)\right) \text{.}
\end{equation}
The final optimization objective is succinctly captured as:
\begin{equation}
\label{eq:final_opt_obj}
\mathcal{L} = \mathcal{L}_{CE} + \mathcal{L}_{GPL} + \mathcal{L}_{DPCL} \text{.}
\end{equation}
Our FDIL process is outlined in Algorithm~\ref{alg:rfdil}.

\begin{algorithm}[ht]\small
    \caption{Rehearsal-free federated domain-incremental learning}
    \label{alg:rfdil}
    \SetKwInOut{Input}{Input}\SetKwInOut{Output}{Output}
    \SetKwRepeat{Do}{do}{while}
    \SetKwProg{procedureA}{Central Server}{}{}
    \SetKwProg{procedureB}{Participants}{}{}
    \Input{Total tasks $T$, Global communication round $R$, client local epoch $E$, number of participants $M$, participant local data $\mathbf{D}_m(x,y)$, participant local model $\theta_m$, learning rate $\eta$}
    \Output{Trained global model $\theta^R$}

    \procedureA{}{
        Initialise global model $\theta^{r}$
        
        \For{$t = 1 $ \KwTo $T$}{
            \For{$r = 1 $ \KwTo $R$}{
                Participants Random Selection 
                
                Broadcast $\theta^{r,t}, \hat{P}^{g}$ $\rightarrow$ \textbf{Selected Participants}
                
                $\theta^r_m$, $LPG^k_m \leftarrow$ \textbf{Selected Participants}
                
                $\theta^{r+1} \leftarrow \sum_{m=1}^N \frac{|\mathbf{D}^t_m|}{|\mathbf{D}^t|} \theta_m^r$
    
                $\hat{P}^{g,k}=\left\{LPG_m\right\}_{m=1}^C  \xrightarrow{\Psi} \left\{LPG_m\right\}_{m=1}^N$ Eq.\ref{eq:global_clustering} \algcomm{Global prompt clustering}

                $\hat{P}^g = \left[\hat{P}^{g,1},\dots, \hat{P}^{g,k}\right]$
            }
            
            $\mathcal{U}_n, \mathcal{U}_b, \mathcal{U}_o \leftarrow$ $M$Sampling \algcomm{Participant task transfer sampling and increment} 
        }
    }

    \procedureB{}{
        $P_m = \left\{ \right\}$ \algcomm{Initialise local prompts collection}
        
        $\theta^{r,t}, \hat{P}^{g}$ $\leftarrow$ \textbf{Central Server} 

        $\bar{P}^g \xleftarrow{\text{Eq.\ref{eq:global_p_avg}}} \hat{P}^{g}$

        \If{$m$ in $\mathcal{U}_b$}{
            $\mathbf{D}^t_m = \text{concat}(\mathbf{D}^{t-1}_m, \mathbf{D}^t_m)$
        }
            
        \For{$e = 0,..., E $}{
            \For{$x_i,y_i \in \mathbf{D}^t_m$}{
                $u_i, p_l, \xi_l, \xi_g = \theta^{r,t}(x_i, \bar{P}^g)$
                
                $\mathcal{L}_{CE} \leftarrow $ Eq.\ref{eq:ce_loss} $(u_i, \xi_l)$
                
                $\mathcal{L}_{GPL} \leftarrow $ Eq.\ref{eq:gpl_loss} $(u_i, \xi_g)$

                $\mathcal{L}_{DPCL} \leftarrow $ Eq.\ref{eq:dpcl_loss} $(p_l, \hat{P}^{g})$

                
                $\mathcal{L} = \mathcal{L}_{CE} + \mathcal{L}_{GPL} + \mathcal{L}_{DPCL}$ 
                
                $\theta^r_m \leftarrow \theta^r_m - \eta \nabla \mathcal{L}(\theta^r_m ; b)$
                
                \If{$e = (E-1)$}{
                  $P_m[k]$ append $p_i$
                }
            }
        }

        \For{$k$ in $P_m$ keys}{ \algcomm{Average local prompts into a representative prompt}
            $LPG_m=\left[\frac{1}{N} \sum_{j=1}^N P^1_{m,j} ,\dots, \frac{1}{N} \sum_{j=1}^N P^k_{m,j} \right]$
        }

        Send $\theta^r_m$, $LPG_m$ $\rightarrow$ \textbf{Central Server}
    }
\end{algorithm}

\section{Experiments}
\subsection{Experimental Setup}
\noindent{\textbf{Datasets.}}
We evaluate our framework on four widely used datasets for image classification tasks. The first, Digits-five~\cite{schrod2023fact}, consists of 215,695 images across 10 classes and five domains, each image sized at 32x32 pixels. The second dataset, OfficeCaltech10~\cite{wang2018visual}, contains 2,533 images distributed over 10 classes and four domains, also in 32x32 pixels. The third, PACS~\cite{li2017deeper}, includes 9,991 images across seven classes and four domains, each image being 224x224 pixels. Lastly, we use a subset of DomainNet~\cite{peng2019moment}, which we have named FedDomainNet, featuring 100,361 images in 48 classes across six domains, with a resolution of 224x224 pixels. 

\noindent{\textbf{Baselines.}} In our study, we benchmark RefFiL against well-known rehearsal-free methods. Specifically, we compare it with the finetune approach, which involves straightforward model updates but is significantly impacted by catastrophic forgetting. Additionally, we consider four prominent rehearsal-free approaches: LwF~\cite{li2017learning}, EWC~\cite{kirkpatrick2017overcoming}, L2P~\cite{wang2022learning} and DualPrompt~\cite{wang2022dualprompt}. These methods are fundamentally based on regularization methods. 

However, they were originally proposed for centralized continual learning contexts, and consider to FDIL is an under-explored field. We adapted these methodologies for our FDIL setting, resulting in variants we refer to as FedLwF, FedEWC, FedL2P and FedDualPrompt, respectively. Particularly, to ensure a fair comparison, we initially deactivated the prompt pool feature in FedL2P and FedDualPrompt. Moreover, to demonstrate the superior efficacy of our RefFiL approach, even when competing methods employ prompt pool as a rehearsal technique, we also evaluate versions of FedL2p and FedDualPrompt with the prompt pool feature reactivated. These special versions are indicated with a $^{\dag}$ superscript in the Table~\ref{tab:res_overall}~\ref{tab:res_combined}~\ref{tab:res_overall_new_order}~\ref{tab:res_combined_new_order}.
\noindent{\textbf{Evaluation Metrics.}} Our evaluation primarily includes top-1 accuracy, reflecting the model's most probable correct prediction. We also report \textit{Average (Avg \%)} accuracy, as defined in iCaRL~\cite{rebuffi2017icarl}, which is the mean accuracy across all learning steps, indicating the model's stability and consistency. Additionally, we present the \textit{Final (Last \%)} accuracy, measured after the last learning step, to assess the model's knowledge retention and consolidation capabilities. Additionally, we consider \textit{Forgetting (FGT)}, which quantifies the average amount of information lost across tasks, and \textit{Backward Transfer (BwT)}, which evaluates how learning new tasks influences the performance of previously learned tasks.

\begin{table*}[th]
  \centering
  \scalebox{0.81}{
    \begin{tabular}{
    l
    c
    c
    c
    c
    c
    c
    c
    c
    c
    c
    c
    c
    c
    c
    c
    c
    }
    \toprule
     & \multicolumn{4}{c}{Digits-five} & \multicolumn{4}{c}{OfficeCaltech10} & \multicolumn{4}{c}{PACS} & \multicolumn{4}{c}{FedDomainNet} \\
    \cmidrule(lr){2-5} \cmidrule(lr){6-9} \cmidrule(lr){10-13} \cmidrule(lr){14-17} 
    Methods & {Avg} & {$\Delta$} & {Last} & {$\Delta$} & {Avg} & {$\Delta$} & {Last} & {$\Delta$} & {Avg} & {$\Delta$} & {Last} & {$\Delta$} & {Avg} & {$\Delta$} & {Last} & {$\Delta$} \\
    \midrule
    Finetune & {77.39} & {9.55\customgreenuparrow} & {49.80} & {12.31\customgreenuparrow} & {44.56} & {9.00\customgreenuparrow}& {19.29} & {14.37\customgreenuparrow}& {40.18}& {15.14\customgreenuparrow}& {30.82}& {13.45\customgreenuparrow}& {28.46} & {0.47\customgreenuparrow} & {18.07} & {0.82\customgreenuparrow} \\
    FedLwF & {77.58} & {9.36\customgreenuparrow} & {56.86} & {5.25\customgreenuparrow} & {46.78} & {6.78\customgreenuparrow}& {28.74} & {4.92\customgreenuparrow}& {40.12}& {15.20\customgreenuparrow}& {26.61}& {17.66\customgreenuparrow}& {27.95} & {0.98\customgreenuparrow} & {17.96} & {1.02\customgreenuparrow} \\
    FedEWC & {78.20} & {8.74\customgreenuparrow} & {45.89} & {16.22\customgreenuparrow} & {44.38} & {9.18\customgreenuparrow}& {15.55} & {18.11\customgreenuparrow}& {40.27}& {15.05\customgreenuparrow}& {27.36}& {16.91\customgreenuparrow}& {26.10} & {2.83\customgreenuparrow} & {18.37} & {0.58\customgreenuparrow} \\
    FedL2P & {83.45} & {3.49\customgreenuparrow} & {57.65} & {4.46\customgreenuparrow} & {46.51} & {7.05\customgreenuparrow}& {26.57} & {7.09\customgreenuparrow}& {49.68}& {5.64\customgreenuparrow}& {35.32}& {8.95\customgreenuparrow}& {25.26} & {3.67\customgreenuparrow} & {18.42} & {0.56\customgreenuparrow} \\
    FedL2P$^{\dag}$ & {84.86} & {2.08\customgreenuparrow} & {60.17 } & {1.94\customgreenuparrow} & {45.41} & {8.15\customgreenuparrow}& {25.20} & {8.46\customgreenuparrow}& {50.00}& {5.32\customgreenuparrow}& {34.52}& {9.75\customgreenuparrow}& {22.18} & {6.75\customgreenuparrow} & {15.59} & {3.39\customgreenuparrow} \\
    FedDualPrompt & {85.15} & {1.79\customgreenuparrow} & {59.30} & {2.81\customgreenuparrow} & {45.15} & {8.41\customgreenuparrow}& {23.82} & {9.84\customgreenuparrow}& {54.05}& {1.27\customgreenuparrow}& {41.07}& {3.20\customgreenuparrow}& {28.25} & {0.68\customgreenuparrow} & {18.05} & {0.93\customgreenuparrow} \\
    FedDualPrompt$^{\dag}$ & {84.39} & {2.55\customgreenuparrow} & {58.34} & {3.77\customgreenuparrow} & {47.86} & {5.70\customgreenuparrow}& {27.76} & {5.90\customgreenuparrow}& {52.79}& {2.53\customgreenuparrow}& {37.62}& {6.65\customgreenuparrow}& {28.53} & {0.40\customgreenuparrow} & {17.76} & {1.22\customgreenuparrow} \\
    
    \midrule
    \rowcolor{ourwork_clr!30} RefFiL & {\textbf{86.94}} & {-} & {\textbf{62.11}} & {-} & {\textbf{53.56}} & {-} & {\textbf{33.66}} & {-} & {\textbf{55.32}}& {-} & {\textbf{44.27}}& {-} & {\textbf{28.93}} & {-} & {\textbf{18.98}} & {-} \\
    \bottomrule
  \end{tabular}
  }
  \caption{Summarised results on four datasets: Digits-five with five incremental tasks, FedDomainNet with six, and both OfficeCaltech10 and PACS with four tasks. Displaying average accuracy (Avg\%) and final task accuracy (Last\%). The amount of change in our method compared to other methods is marked as $\Delta$, 
  The results of our RefFiL are highlighted in colour \temtextclr{0.2cm} \\
  $^{\dag}$ Means this baseline enabled prompt pool
  }
  \label{tab:res_overall}
\end{table*}

\begin{table*}[th]
  \centering
  \scalebox{0.84}{
    \begin{tabular}{
    l
    c
    c
    c
    c
    c
    c
    c
    c
    c
    c
    c
    c
    c
    c
    c
    c
    }
    \toprule
     & \multicolumn{4}{c}{Digits-five} & \multicolumn{4}{c}{OfficeCaltech10} & \multicolumn{4}{c}{PACS} & \multicolumn{4}{c}{FedDomainNet} \\
    \cmidrule(lr){2-5} \cmidrule(lr){6-9} \cmidrule(lr){10-13} \cmidrule(lr){14-17} 
    Methods & {Avg} & {$\Delta$} & {Last} & {$\Delta$} & {Avg} & {$\Delta$} & {Last} & {$\Delta$} & {Avg} & {$\Delta$} & {Last} & {$\Delta$} & {Avg} & {$\Delta$} & {Last} & {$\Delta$} \\
    \midrule
    Finetune & {59.84}& {9.52\customgreenuparrow}& {58.20}& {2.64\customgreenuparrow}& {37.60}& {6.73\customgreenuparrow}& {25.20}& {13.19\customgreenuparrow}& {46.99}& {4.09\customgreenuparrow}& {38.97}& {7.75\customgreenuparrow}& {31.85}& {1.49\customgreenuparrow}& {11.58}& {4.16\customgreenuparrow}\\
    FedLwF & {65.22}& {4.14\customgreenuparrow}& {59.36}& {1.48\customgreenuparrow}& {38.76}& {5.57\customgreenuparrow}& {25.20}& {13.19\customgreenuparrow}& {43.43}& {7.65\customgreenuparrow}& {30.17}& {16.55\customgreenuparrow}& {31.33}& {2.01\customgreenuparrow}& {11.01}& {4.73\customgreenuparrow}\\
    FedEWC & {64.00}& {5.36\customgreenuparrow}& {59.54}& {1.30\customgreenuparrow}& {38.26}& {6.07\customgreenuparrow}& {27.95}& {10.44\customgreenuparrow}& {43.60}& {7.48\customgreenuparrow}& {30.22}& {16.50\customgreenuparrow}& {30.38}& {2.96\customgreenuparrow}& {12.03}& {3.71\customgreenuparrow}\\
    FedL2P & {66.00}& {3.36\customgreenuparrow}& {59.84}& {1.00\customgreenuparrow}& {41.58}& {2.75\customgreenuparrow}& {34.45}& {3.94\customgreenuparrow}& {45.99}& {5.09\customgreenuparrow}& {31.02}& {15.70\customgreenuparrow}& {25.19}& {8.15\customgreenuparrow}& {9.51}& {6.23\customgreenuparrow}\\
    FedL2P$^{\dag}$ & {64.45}& {4.91\customgreenuparrow}& {59.74}& {1.10\customgreenuparrow}& {41.24}& {3.13\customgreenuparrow}& {31.50}& {6.89\customgreenuparrow}& {45.39}& {5.69\customgreenuparrow}& {35.42}& {11.30\customgreenuparrow}& {22.95}& {10.39\customgreenuparrow}& {7.32}& {8.42\customgreenuparrow}\\
    FedDualPrompt & {65.31}& {4.05\customgreenuparrow}& \textbf{{60.94}} & {0.10\customreddownarrow}& {40.47}& {3.86\customgreenuparrow}& {31.50}& {6.89\customgreenuparrow}& {48.41}& {2.67\customgreenuparrow}& {42.32}& {4.40\customgreenuparrow}& {33.09}& {0.25\customgreenuparrow}& {14.54}& {1.20\customgreenuparrow}\\
    FedDualPrompt$^{\dag}$ & {66.61}& {2.75\customgreenuparrow}& {\textbf{60.94}}& {0.10\customreddownarrow}& {39.73}& {4.6\customgreenuparrow}& {30.91}& {7.48\customgreenuparrow}& {47.64}& {3.44\customgreenuparrow}& {42.82}& {3.90\customgreenuparrow}& {30.11}& {3.23\customgreenuparrow}& {14.54}& {1.20\customgreenuparrow}\\
    \midrule
    \rowcolor{ourwork_clr!30} RefFiL & {\textbf{69.36}}& {-} & {60.84}& {-} & {\textbf{44.33}}& {-} & {\textbf{38.39}}& {-} & {\textbf{51.08}}& {-} & {\textbf{46.72}}& {-} & {\textbf{33.34}}& {-} & {\textbf{15.74}}& {-} \\
    \bottomrule
  \end{tabular}
  }
  \caption{Summarised results in new domain order of Table~\ref{tab:res_overall}. 
  }
  \vspace{-0.3cm}
  \label{tab:res_overall_new_order}
\end{table*}

\noindent{\textbf{Implementation details.}}
In our study, all methods are implemented using PyTorch and run on a single NVIDIA RTX 4090 GPU, using
ResNet10 as the feature extractor classification model backbone. We designed a simple embedding model as the feature map tokenizer, similar to ViT~\cite{dosovitskiy2020image}, with initialized-only and frozen parameters for feature embedding. The FedAvg~\cite{mcmahan2017communication} algorithm was employed for global model aggregation. Each task involves training the model over $R = 30$ rounds. The number of epochs for local updates on each client is set at 20. We use a learning rate of 0.04 for FedDomainNet, 0.06 for OfficeCaltech10 and 0.03 for the rest of the datasets to get the proper performance. Additionally, we use SGD as an optimizer for all experiments. Unless otherwise specified, the constraint factor $\lambda$ in the EWC method is fixed at 300, and the temperature for distillation is set to the default value of 2. 
For the experiments on the \textbf{Digit-Five}, \textbf{PACS}, and \textbf{FedDomainNet} datasets, as shown in Tables~\ref{tab:res_overall}, \ref{tab:res_overall_new_order}, \ref{tab:res_combined}, \ref{tab:res_combined_new_order}, and \ref{table:hyparam_ana}, the setup starts with 20 clients, of which 10 are initially selected, with 2 additional clients incrementally introduced per new task. For the \textbf{OfficeCaltech10} dataset, considering its data sample limitations, the setup begins with 10 clients, of which 5 are initially selected, and 1 additional client is incrementally introduced per new task. Since this work focuses on domain-incremental learning, after the first task, all clients retain the same number of classes but represent distinct domains, and for each task $t$, 80\% of the $M$ clients from task $t$ transition to the next task. In the experiments summarized in Table~\ref{tab:digits_diff_sel}, the setup includes 10 clients initialized at the start, with 1 additional client joining per task increment, while for the results in Table~\ref{tab:officecaltech10_diff_sel}, the \textbf{Digits Dataset} is starting with 20 clients and incrementing by 2 clients per task. The hyperparameters in Eq.~\ref{eq:tem_decay} are fixed as follows: $\tau_{\text{min}} = 0.3$, $\gamma = 0.1$, $\beta = 0.05$, and $\tau = 0.9$.

\begin{table*}[th]
  \centering
  \scalebox{0.87}{
    \begin{tabular}{
      l
      c
      c
      c
      c
      c
      c
      c
      ||
      c
      c
      c
      c
      c
    }
    \toprule
    & \multicolumn{6}{c}{Task 1 $\rightarrow$ 5 on Digit-Five} & & \multicolumn{4}{c}{Task 1 $\rightarrow$ 4 on OfficeCaltech10} & \\ 
    \cmidrule(lr){2-7} \cmidrule(lr){9-12}
    Methods & MNIST & MNIST-M & USPS & SVHN & SYN & -- & Avg & Amazon & Caltech & Webcam & DSLR & Avg \\
    \midrule
    Finetune & {99.68} & {97.75} & {63.87} & {75.84} & {49.80}& -- & {77.39} & {76.56} & {57.79} & {24.58} & {19.29} & {44.56} \\
    FedLwF & {99.68} & {92.80} & {69.16} & {69.39} & {56.86}& -- & {77.58} & {76.56} & {53.24} & {28.57} & {28.74} & {46.78} \\
    FedEWC & {99.68} & {97.48} & {74.63} & {73.32} & {45.89} & --& {78.20} & {76.56} & {56.59} & {29.83} & {15.55} & {44.38} \\
    FedL2P & {99.66} & {98.06} & {80.01} & {81.89} & {57.65} & --& {83.45} & {76.56} & {51.80} & {31.09} & {26.57} & {46.51} \\
    FedL2P$^{\dag}$ & {99.64} & {97.65} & {85.18} & {81.65} & {60.17} & --& {84.86} & {71.35} & {55.88} & {29.20} & {25.20} & {45.41} \\
    FedDualPrompt & {99.67} & {97.96} & {86.88} & {81.95} & {59.30} & --& {85.15} & {74.48} & {50.36} & {31.93} & {23.82} & {45.15} \\
    FedDualPrompt$^{\dag}$ & {99.65} & {97.90} & {84.68} & {81.40} & {58.34} & --& {84.39} & {75.90} & {53.96} & {33.82} & {27.76} & {47.86} \\
    \midrule
    \rowcolor{ourwork_clr!30} RefFiL & {\textbf{99.68}} & {\textbf{98.25}} & {\textbf{90.96}} & {\textbf{83.70}} & {\textbf{62.11}}& -- & {\textbf{86.94}}  & {\textbf{78.65}} & {\textbf{61.15}} & {\textbf{40.76}} & {\textbf{33.66}} & {\textbf{53.56}} \\
    \bottomrule
    \toprule
    & \multicolumn{6}{c}{Task 1 $\rightarrow$ 6 on FedDomainNet} & & \multicolumn{4}{c}{Task 1 $\rightarrow$ 4 on PACS} & \\ 
    \cmidrule(lr){2-7} \cmidrule(lr){9-12}
    Methods & Clipart & Infograph & Painting & Quickdraw & Real & Sketch & Avg & Photo & Cartoon & Sketch & Art Painting & Avg \\
    \midrule
    Finetune & {\textbf{51.48}} & {15.89} & {28.05} & {\textbf{27.84}}  & {29.45} & {18.07} & {28.46} & {61.68} & {47.45} & {36.12} & {30.82} & {40.18} \\
    FedLwF & {\textbf{51.48}} & {18.10} & {26.71} & {25.98}  & {27.47} & {17.96} & {27.95} & {61.68} & {47.07} & {25.11} & {26.61} & {40.12} \\
    FedEWC & {50.76} & {15.46} & {22.66} & {21.87}  & {27.45} & {18.37} & {26.10} & {63.17} & {47.70} & {23.66} & {27.36} & {40.27} \\
    FedL2P & {40.55} & {13.19} & {21.09} & {28.15}  & {30.13} & {18.42} & {25.26} & {64.97} & {48.32} & {\textbf{50.09}} & {35.32} & {49.68} \\
    FedL2P$^{\dag}$ & {37.63} & {9.29} & {16.79} & {27.09}  & {26.68} & {15.59} & {22.18} & {65.57} & {54.67} & {45.25} & {34.52} & {50.00} \\
    FedDualPrompt & {51.17} & {19.48} & {28.74} & {22.68}  & {29.40} & {18.05} & {28.25} & {73.65} & {56.54} & {44.93} & {41.07} & {54.05} \\
    FedDualPrompt$^{\dag}$ & {51.14} & {20.20} & {28.91} & {23.09}  & {30.07} & {17.76} & {28.53} & {\textbf{75.75}} & {54.55} & {43.23} & {37.62} & {52.79} \\
    \midrule
    \rowcolor{ourwork_clr!30} RefFiL & {51.27} & {\textbf{20.91}} & {\textbf{29.23}} & {22.57} & {\textbf{30.62}} & {\textbf{18.98}} & {\textbf{28.93}} & {73.95} & {\textbf{59.90}} & {43.17} & {\textbf{44.27}} & {\textbf{55.32}} \\
    \bottomrule
    \end{tabular}
  }
  \caption{Comparison of RefFiL's performance with five baseline methods on four widely used datasets, showcasing average accuracy (Avg \%) and accuracy for each domain task (\%). At each task step, all clients have an equal number of classes but \textit{quantity shift} data.} 
  \label{tab:res_combined}
\end{table*}

\begin{table*}[th]
  \centering
  \scalebox{0.87}{
    \begin{tabular}{
      l
      c
      c
      c
      c
      c
      c
      c
      ||
      c
      c
      c
      c
      c
    }
    \toprule
    & \multicolumn{6}{c}{Task 1 $\rightarrow$ 5 on Digit-Five} & & \multicolumn{4}{c}{Task 1 $\rightarrow$ 4 on OfficeCaltech10} & \\ 
    \cmidrule(lr){2-7} \cmidrule(lr){9-12}
    Methods & SVHN & MNIST & SYN & USPS & MNIST-M & -- & Avg & Caltech  & Amazon & DSLR& Webcam & Avg \\
    \midrule
    Finetune & {94.97} & {58.35} & {49.04} & {38.66} & {58.20} & -- & {59.84} & {49.78} & {58.27} & {17.15} & {25.20} & {37.60} \\
    FedLwF & {94.97} & {73.21} & {54.73} & {43.82} & {59.36} & -- & {65.22} & {49.78} & {57.79} & {22.27} & {25.20} & {38.76} \\
    FedEWC & {95.03} & {64.32} & {50.22} & {50.88} & {59.54} & -- & {64.00} & {48.00} & {56.83} & {20.27} & {27.95} & {38.26} \\
    FedL2P & {94.85} & {73.54} & {53.19} & {48.56} & {59.84} & -- & {66.00} & {49.78} & {58.03} & {24.05} & {34.45} & {41.58} \\
    FedL2P$^{\dag}$ & {94.80} & {73.45} & {51.07} & {43.21} & {59.74} & -- & {64.45} & {50.67} & {58.27} & {\textbf{24.50}} & {31.50} & {41.24} \\
    FedDualPrompt & {94.78} & {70.71} & {54.06} & {46.04} & {\textbf{60.94}} & -- & {65.31} & {48.00} & {58.75} & {23.61} & {31.50} & {40.47} \\
    FedDualPrompt$^{\dag}$ & {94.65} & {77.02} & {54.43} & {46.01} & {\textbf{60.94}} & -- & {66.61} & {50.22} & {57.07} & {20.71} & {30.91} & {39.73} \\
    \midrule
    \rowcolor{ourwork_clr!30} RefFiL & {\textbf{95.35}} & {\textbf{76.03}} & {\textbf{59.90}} & {\textbf{54.68}} & {60.84} & -- & {\textbf{69.36}}  & {\textbf{52.00}} & {\textbf{63.31}} & {23.61} & {\textbf{38.39}} & {\textbf{44.33}} \\
    \bottomrule
    \toprule
    & \multicolumn{6}{c}{Task 1 $\rightarrow$ 6 on FedDomainNet} & & \multicolumn{4}{c}{Task 1 $\rightarrow$ 4 on PACS} & \\ 
    \cmidrule(lr){2-7} \cmidrule(lr){9-12}
    Methods & Infograph & Sketch & Quickdraw & Real & Painting & Clipart & Avg & Cartoon & Photo & Sketch & Art Painting & Avg \\
    \midrule
    Finetune & {68.84} & {33.94} & {\textbf{28.94}} & {26.12}  & {21.73} & {11.58} & {31.85} & {68.23} & {40.97} & {39.77} & {38.97} & {46.99} \\
    FedLwF & {68.84} & {34.87} & {28.82} & {23.88} & {20.53} & {11.01} & {31.33} & {68.23} & {36.11} & {39.21} & {30.17} & {43.43} \\
    FedEWC & {68.11} & {34.66} & {24.63} & {24.10}  & {18.75} & {12.03} & {30.38} & {69.94} & {38.23} & {36.00} & {30.22} & {43.60} \\
    FedL2P & {53.39} & {26.76} & {27.57} & {17.92} & {15.98} & {9.51} & {25.19} & {68.23} & {42.34} & {\textbf{42.73}} & {31.02} & {45.99} \\
    FedL2P$^{\dag}$ & {51.89} & {24.86} & {26.37} & {14.64} & {12.62} & {7.32} & {22.95} & {66.95} & {44.71} & {34.49} & {35.42} & {45.39} \\
    FedDualPrompt & {\textbf{70.67}} & {41.50} & {25.82} & {25.30}  & {20.73} & {14.54} & {33.09} & {69.94} & {41.34} & {40.03} & {42.32} & {48.41} \\
    FedDualPrompt$^{\dag}$ & {70.33} & {40.63} & {25.69} & {25.48}  & {21.40} & {14.54} & {30.11} & {66.74} & {41.72} & {39.27} & {42.82} & {47.64} \\
    \midrule
    \rowcolor{ourwork_clr!30} RefFiL & {69.02} & {\textbf{42.48}} & {24.70} & {\textbf{26.19}} & {\textbf{21.93}} & {\textbf{15.74}} & {\textbf{33.34}} & {\textbf{73.13}} & {\textbf{45.33}} & {39.14} & {\textbf{46.72}} & {\textbf{51.08}} \\
    \bottomrule
    \end{tabular}
  }
  \caption{Comparison of RefFiL's performance in new domain order of table~\ref{tab:res_combined}.}  
  \label{tab:res_combined_new_order}
\end{table*}

\begin{table*}[htbp]
\centering
\small 
\scalebox{0.78}{
\begin{tabular}{lcccccccccccccccc}
\toprule
 & \multicolumn{4}{c}{\textbf{Sel 8, 80\% of $M$}} & 
 \multicolumn{4}{c}{\textbf{Sel 2, 80\% of $M$}} & 
 \multicolumn{4}{c}{\textbf{Sel 5, 50\% of $M$}} & 
 \multicolumn{4}{c}{\textbf{Sel 5, 90\% of $M$}} \\
\cmidrule(lr){2-5} \cmidrule(lr){6-9} \cmidrule(lr){10-13} \cmidrule(lr){14-17}
\textbf{Method} & \textbf{AVG} & \textbf{Last} & \textbf{FGT} & \textbf{BwT} & 
\textbf{AVG} & \textbf{Last} & \textbf{FGT} & \textbf{BwT} & 
\textbf{AVG} & \textbf{Last} & \textbf{FGT} & \textbf{BwT} & 
\textbf{AVG} & \textbf{Last} & \textbf{FGT} & \textbf{BwT} \\
\midrule
Finetune         & 47.01 & 20.28 & 0.348 & -0.290 & 47.06 & 25.79 & 0.287 & -0.288 & 48.24 & 25.59 & 0.294 & -0.252 & 47.39 & 25.00 & 0.313 & -0.267 \\
FedLwF           & 46.19 & 23.23 & 0.342 & -0.286 & \textbf{49.22}& 29.53 & 0.262 & -0.202 & 48.04 & 26.18 & 0.281 & -0.228 & 47.81 & 26.38 & 0.288 & -0.227 \\
FedEWC           & 45.15 & 18.50 & 0.326 & -0.259 & 45.26 & 25.20 & \textbf{0.234}& -0.197 & 47.83 & 25.00 & 0.292 & -0.253 & 47.64 & 22.24 & 0.315 & -0.255 \\
FedL2P           & 50.04 & 23.03 & 0.301 & -0.222 & 48.50 & 29.53 & 0.265 & -0.215 & 51.20 & 28.74 & 0.291 & -0.232 & 51.41 & 27.36 & 0.311 & -0.231 \\
FedL2P$^{\dag}$         & 50.93 & 25.20 & 0.323 & -0.243 & 47.76 & 28.15 & 0.259 & -0.209 & 53.38 & 29.13 & 0.282 & -0.211 & 51.86 & 27.95 & 0.301 & -0.234 \\
FedDualPrompt    & 50.45 & 27.95 & 0.296 & -0.232 & 47.04 & 25.98 & 0.263 & -0.199 & 50.65 & 28.74 & 0.307 & -0.237 & 50.44 & 29.33 & 0.305 & -0.246 \\
FedDualPrompt$^{\dag}$  & 50.06 & 25.00 & 0.321 & -0.252 & 47.53 & 27.95 & 0.279 & -0.217 & 51.97 & 27.56 & 0.309 & -0.231 & 52.20 & 29.13 & 0.295 & -0.219 \\
\rowcolor{ourwork_clr!30} RefFiL  & \textbf{54.59}& \textbf{33.86}& \textbf{0.278}& \textbf{-0.198}& \textbf{49.22}& \textbf{30.12}& 0.237 & \textbf{-0.192}& \textbf{54.74}& \textbf{34.25}& \textbf{0.270}& \textbf{-0.198}& \textbf{54.22}& \textbf{31.89}& \textbf{0.274}& \textbf{-0.200}\\
\bottomrule
\end{tabular}%
}
\caption{Results on the OfficeCaltech10 Dataset, following the same domain order as in Table~\ref{tab:res_combined}.}
\label{tab:officecaltech10_diff_sel}
\end{table*}

\begin{table}[htbp]
\centering
\small 
\begin{tabular}{lcccc}
\toprule
\multicolumn{1}{l}{} & 
\multicolumn{4}{c}{\textbf{Sel 10, 90\% of $M$}} \\
\cmidrule(lr){2-5}
\textbf{Method} & \textbf{AVG} & \textbf{Last} & \textbf{FGT} & \textbf{BwT} \\
\midrule
Finetune         & 79.80 & 53.49 & 0.405 & -0.256 \\
FedLwF           & 78.64 & 53.51 & 0.443 & -0.276 \\
FedEWC           & 78.30 & 51.98 & 0.317 & -0.226 \\
FedL2P           & 84.00 & 57.03 & 0.374 & -0.214 \\
FedL2P$^{\dag}$  & 85.42 & 58.88 & 0.323 & -0.191 \\
FedDualPrompt    & 84.57 & 58.06 & 0.331 & -0.201 \\
FedDualPrompt$^{\dag}$ & 84.70 & 57.71 & 0.324 & -0.193 \\
\rowcolor{ourwork_clr!30} RefFiL (Our)   & \textbf{86.84}& \textbf{62.50}& \textbf{0.313}& \textbf{-0.182}\\
\bottomrule
\end{tabular}%
\caption{Results on the Digits Dataset, following the same domain order as in Table~\ref{tab:res_combined}.} 
\label{tab:digits_diff_sel}
\end{table}

\subsection{Qualitative Results}
In our study, we evaluated the performance of seven baseline methods alongside our proposed method, RefFiL, across four widely used datasets: Digit-Five, OfficeCaltech10, FedDomainNet, and PACS. The results, detailed in Table~\ref{tab:res_overall}, Table~\ref{tab:res_combined}, Table~\ref{tab:res_overall_new_order}, and Table~\ref{tab:res_combined_new_order}, indicate that RefFiL consistently outperforms the baseline methods in both average and last task accuracy, even under the impact of catastrophic forgetting. Notably, RefFiL maintains its advanced performance, regardless of changes in domain order. Furthermore, to provide a comprehensive comparison across different settings and evaluation metrics, we conducted three additional experiments, detailed in Tables~\ref{tab:digits_diff_sel} and \ref{tab:officecaltech10_diff_sel}, which further exhibit the outstanding performance of RefFiL. 

\noindent \textbf{RefFiL efficiently mitigates forgetting.}
We report detailed performance in Table~\ref{tab:res_combined}. The $^{\dag}$ indicates baselines enabled with a prompt pool to save generated prompts, which are sampled to better mitigate forgetting. The summarized results in Table~\ref{tab:res_overall} demonstrate that RefFiL consistently achieves state-of-the-art (SOTA) performance across all four datasets in terms of both average accuracy (Avg) and final task accuracy (Last). For example, in smaller and less complex datasets such as Digit-Five, RefFiL significantly outperforms other baselines, achieving improvements in Avg by up to 9.55\% and in Last by up to 16.22\%. Similarly, in OfficeCaltech10, RefFiL exceeds baselines by as much as 9.18\% in Avg and 18.11\% in Last, showcasing its ability to effectively manage diverse domains. In larger, more complex datasets like PACS, RefFiL maintains its leadership, achieving Avg and Last improvements of up to 15.20\% and 17.66\%, respectively. Even under the challenging conditions of FedDomainNet, characterized by sparse data distributions across 48 classes, RefFiL demonstrates robust performance, improving Avg by up to 6.75\% and Last by up to 3.39\% compared to baselines.

To further evaluate RefFiL’s adaptability under diverse configurations, we conducted additional experiments, as detailed in Table~\ref{tab:officecaltech10_diff_sel} for OfficeCaltech10 and Table~\ref{tab:digits_diff_sel} for Digit-Five. These experiments explore RefFiL’s performance under varying client selection percentages and task transfer configurations. In OfficeCaltech10, RefFiL achieves consistently superior results, with Avg reaching 54.59\% and Last 33.86\% under the select 8 clients, 80\% task transfer setup. RefFiL also demonstrates the lowest FGT at 0.278 and the most favorable BwT at -0.198 in this setup, reflecting its exceptional capacity to retain knowledge from earlier tasks while effectively incorporating new learning. Compared to baselines such as FedLwF (FGT: 0.342, BwT: -0.286) and FedL2P (FGT: 0.301, BwT: -0.222), RefFiL significantly reduces forgetting and enhances backward transfer, achieving a more optimal balance between forward and backward learning. These results validate RefFiL’s ability to mitigate catastrophic forgetting while ensuring stable performance across diverse configurations.

Similarly, in Digit-Five, RefFiL effectively adapts to domain selection sizes, achieving 86.84\% Avg and 62.50\% Last in the select 10 clients, 90\% task transfer setup. It maintains competitive forgetting (FGT: 0.313) and achieves the most favorable backward transfer (BwT: -0.182), further reinforcing its capacity to integrate new knowledge while minimizing performance degradation on prior tasks. These results underscore RefFiL’s adaptability and robustness across different scenarios, highlighting its ability to consistently deliver high performance and maintain stability in diverse FDIL tasks.

\noindent \textbf{RefFiL produces domain order-independent consistency results.}
To demonstrate the consistency of our results across different domain introduction orders, we generated new results presented in Table~\ref{tab:res_combined_new_order} and summarized in Table~\ref{tab:res_overall_new_order}. In these experiments, we randomly altered the order of domain introductions. Specifically, in Table~\ref{tab:res_combined_new_order}, the domain orders for Digit-Five, OfficeCaltech10, and FedDomainNet were significantly disrupted, while in PACS, only the order of the first two domains was changed.

\begin{table}[t]
\centering
    \scalebox{0.86}{
    \begin{tabular}{c|c|c|c|c|c|c|c}
    \toprule
    \multicolumn{1}{c|}{\textbf{Methods}} & \textbf{CDAP} & \textbf{GPL} & \textbf{DPCL} & \textbf{Avg} & $\Delta$ & \textbf{Last} & $\Delta$ \\
    \midrule
     &  &  &  & 44.56 & - & 19.29 & - \\
    \cmidrule{2-8}
     & \checkmark &  &  & 49.78 & 5.22\customgreenuparrow & 27.56 & 8.27\customgreenuparrow \\
    \cmidrule{2-8}
    RefFiL &  & \checkmark &  & 47.94 & 3.38\customgreenuparrow & 26.38 & 7.09\customgreenuparrow \\
    \cmidrule{2-8}
     & \checkmark & \checkmark &  & 50.32 & 5.76\customgreenuparrow & 25.39 & 6.10\customgreenuparrow \\
    \cmidrule{2-8}
    &  & \checkmark & \checkmark & 49.45 & 4.89\customgreenuparrow & 30.12 & 10.83\customgreenuparrow \\
    \cmidrule{2-8}
     & \checkmark & \checkmark & \checkmark & \textbf{53.56} & \textbf{9.00}\customgreenuparrow  &  \textbf{33.66} & \textbf{14.37}\customgreenuparrow \\
    \bottomrule
    \end{tabular}
}
\caption{Ablation study of RefFiL on OfficeCaltech10 dataset, and the amount of improvement in RefFiL with different components compared to baseline~(Finetune) is marked as $\Delta$.}
\label{table:ab}
\end{table}

This setup highlights the robustness of our method, showing that RefFiL maintains high performance despite variations in the sequence of domain introductions. Compared to the results in Table~\ref{tab:res_overall}, although FedDualPrompt and FedDualPrompt$^{\dag}$ achieved better performance in the Last metric, RefFiL consistently outperformed these methods in the more critical Avg metric, indicating its superior overall performance. For both the OfficeCaltech10 and PACS datasets, RefFiL's advantages in Avg and Last metrics were magnified compared to most of the baselines. In the most challenging dataset, FedDomainNet, while RefFiL's advantage in Avg reduced from 0.40\% to 0.20\%, its Last metric slightly improved from 0.93\% to 1.20\%. These results significantly boosted RefFiL's advantages over others.

These results illustrate that RefFiL efficiently and stably mitigates catastrophic forgetting by enhancing the model's ability to learn from domain-invariant knowledge. The consistent high performance across various domain orders underscores RefFiL's robustness and adaptability in FDIL scenarios.

\subsection{Ablation Study}
To understand how different components can mitigate catastrophic forgetting problems, we conduct the experiments as shown in Table~\ref{table:ab}; we dissect the impact of each module within RefFiL, as well as their synergistic effects. Each experimental configuration involving a separate component is indicated by a checkmark (\checkmark).

Our findings reveal that using either the CDAP or GPL module individually enhances both average and final task accuracy, but their combination results in only a slight improvement in average accuracy compared to when they are used separately. Considering that the DPCL module cannot function in isolation, we explored its effectiveness in conjunction with the GPL module. This combination markedly improved the model's performance, especially in the accuracy of the final task. Most importantly, when all three core components CDAP, GPL, and DPCL are integrated, we observe a substantial improvement in both average and last-task accuracy.

An important observation is that while combining CDAP with GPL can improve the model’s average performance, it does not outperform the individual use of CDAP or GPL. This may be due to the increased domain data heterogeneity in the final task. The personalized prompts generated by CDAP and shared by GPL might cause difficulty for the model in distinguishing between heterogeneous domain information.

This comprehensive ablation study confirms the essentiality and complementarity of each module. Their collective operation significantly enhances the global model's performance in domain-incremental scenarios. The results validate the necessity of each component, underscoring their collective contribution to achieving optimal outcomes in FDIL.

\section{Hyperparameter sensitivity analysis}
\begin{table}[t]
\centering
\scalebox{0.95}{
\begin{tabular}{c|c|c|c|c|c|c|c}
\toprule
\multicolumn{1}{c|}{\textbf{Exp}} & $\tau$ & $\tau_{\text{min}}$ & $\gamma$ & $\beta$ & $\tau^{\prime}$ (3rd) & \textbf{Avg} & \textbf{Last} \\
\midrule
1 & 0.5 & 0.2 & 0.15 & 0.1 & 0.325 & 42.17 & 36.22 \\
\cmidrule{2-8}
2 & 0.5 & 0.4 & 0.05 & 0.05 & 0.425 & 42.32 & 37.01 \\
\cmidrule{2-8}
3 & 0.7 & 0.3 & 0.1 & 0.05 & 0.560 & 43.14 & 37.80 \\
\cmidrule{2-8}
4 & 0.9 & 0.2 & 0.05 & 0.1 & 0.675 & 42.89 & 36.81 \\
\cmidrule{2-8}
5 & 0.9 & 0.4 & 0.05 & 0.01 & 0.837 & 42.30 & 36.02 \\
\cmidrule{2-8}
w/o $\tau^{\prime}$ & - & - & - & - & - & 41.97 & 34.39 \\
\cmidrule{2-8}
\rowcolor{ourwork_clr!30} Our & 0.9 & 0.3 & 0.1 & 0.05 & 0.720 & 44.33 & 38.39 \\
\bottomrule
\end{tabular}
}
\caption{Experimental configurations for sensitivity analysis of hyperparameters in OfficeCaltech10 dataset, with domain order shown in Table~\ref{tab:res_combined_new_order}.}
\label{table:hyparam_ana}
\end{table}

To evaluate the sensitivity of the hyperparameters in Eq.\ref{eq:tem_decay}, we conducted a grid search. Based on our preliminary experiments, we identified a suitable range for each hyperparameter as follows: $\tau$ in ${0.5, 0.7, 0.9}$, $\tau_{\text{min}}$ in ${0.2, 0.3, 0.4}$, $\gamma$ in ${0.05, 0.1, 0.15}$, and $\beta$ in ${0.01, 0.05, 0.1}$. These parameters can form a total of 81 combinations. Given that the final value of $\tau^{\prime}$ influences the loss function $\mathcal{L}_{DPCL}$, we selected five representative groups of hyperparameter combinations to explore how these variations affect the model’s performance. Additionally, we conducted a comparative experiment without applying the overall decay parameter $\tau^{\prime}$, marked as w/o $\tau^{\prime}$. As illustrated in Table\ref{table:hyparam_ana}, different combinations of these parameters resulted in only minor variations in the model’s performance metrics. Notably, in the last task, where the overall domain diversity increased, $\tau^{\prime}$ effectively mitigated the impact of heterogeneous domain data on model performance.

\section{Conclusion}
\label{sec:app_limit}
In this paper, we introduced RefFiL, a rehearsal-free federated domain-incremental learning framework designed to augment cross-domain prediction robustness by enhancing the acquisition of domain-invariant knowledge. Our approach includes a client-wise domain adaptive prompt generator for creating personalized, instance-level prompts, a global prompt learning paradigm for breaking down domain-specific information silos, and a novel domain-specific prompt contrastive learning method with temperature decay for discriminating between relevant and irrelevant prompts. Through extensive experiments, we show that RefFiL framework excels in capturing domain invariant knowledge mitigating catastrophic forgetting in federated domain-incremental learning. It is also important to acknowledge that while RefFiL is proposed for learning domain-invariant knowledge through prompting, it may also contribute to the development of fine-tuning large pretrained models in domain-specific areas.

\section*{Acknowledgment}
The research is partially supported by National Edge AI Hub for Real Data: Edge Intelligence for Cyber-disturbances and Data Quality, EPSRC, EP/Y028813/1. And UK-Australia Centre in a Secure Internet of Energy: Supporting Electric Vehicle Infrastructure at the “Edge” of the Grid, EPSRC, EP/W003325/1.

\bibliographystyle{plain}
\bibliography{main.bib}

\end{document}